\documentclass{article}

\usepackage{microtype}
\usepackage{graphicx}
\usepackage{subfigure}
\usepackage{booktabs} 

\usepackage{hyperref}

 \usepackage[accepted]{icml2023}

\usepackage{amsmath}
\usepackage{amssymb}
\usepackage{mathtools}
\usepackage{amsthm}

\usepackage{graphicx} 
\usepackage{booktabs}

\usepackage{amsfonts}       
\usepackage{nicefrac}       
\usepackage{microtype}      
\usepackage{xcolor}         
\usepackage{microtype}      
\usepackage{graphicx}
\usepackage{subfigure} 
\usepackage{mathrsfs}
\usepackage{multirow}
\usepackage{float}
\usepackage{setspace}
\usepackage{mathrsfs} 
\usepackage{enumitem} 

\usepackage[capitalize,noabbrev]{cleveref}

\theoremstyle{plain}

\theoremstyle{definition}

\theoremstyle{remark}

\usepackage[textsize=tiny]{todonotes}

\icmltitlerunning{NP-SemiSeg: When Neural Processes meet Semi-Supervised Semantic Segmentation}

\begin{document}

\twocolumn[
\icmltitle{NP-SemiSeg: When Neural Processes meet Semi-Supervised Semantic Segmentation}



\icmlsetsymbol{equal}{*}

\begin{icmlauthorlist}
\icmlauthor{Jianfeng Wang}{yyy}
\icmlauthor{Daniela Massiceti}{comp}
\icmlauthor{Xiaolin Hu}{sch1}
\icmlauthor{Vladimir Pavlovic}{sch2}
\icmlauthor{Thomas Lukasiewicz}{tuw,yyy} 
\end{icmlauthorlist}

\icmlaffiliation{yyy}{Department of Computer Science, University of Oxford, UK.}
\icmlaffiliation{comp}{Microsoft Research, Cambridge, UK.}
\icmlaffiliation{sch1}{Department of Computer Science and Technology, Tsinghua University, Beijing, China.}
\icmlaffiliation{sch2}{Department of Computer Science, Rutgers University, New Jersey, USA.}
\icmlaffiliation{tuw}{Vienna University of Technology, Austria} 

\icmlcorrespondingauthor{Jianfeng Wang}{jianfeng.wang@cs.ox.ac.uk} 

\icmlkeywords{Machine Learning, ICML}

\vskip 0.3in
]



\printAffiliationsAndNotice{}  

\begin{abstract} 
Semi-supervised semantic segmentation involves assigning pixel-wise labels to unlabeled images at training time. This is useful in a wide range of real-world applications where collecting pixel-wise labels is not feasible in  time or cost. Current approaches to semi-supervised semantic segmentation work by predicting pseudo-labels for each pixel from a class-wise probability distribution output by a model. If the predicted probability distribution is incorrect, however, this leads to poor segmentation results, which can have knock-on consequences in safety critical systems, like medical images or self-driving cars. It is, therefore, important to understand what a model does not know, which is mainly achieved by uncertainty quantification. Recently, neural processes (NPs) have been explored in semi-supervised image classification, and they have been a computationally efficient and effective method for uncertainty quantification.  
In this work, we move one step forward by adapting NPs to semi-supervised semantic segmentation, resulting in a new model called NP-SemiSeg.  
We experimentally evaluated NP-SemiSeg on the public benchmarks PASCAL VOC 2012 and Cityscapes, with different training settings, and the  results verify its effectiveness. 
\end{abstract}

\section{Introduction} 

Semi-supervised image segmentation has seen a rapid pro\-gress in recent years and involves assigning class labels to every pixel in an unlabeled image at training time. This has many real-world applications, from medical imaging to autonomous driving systems, where the cost and time to annotate  large-scale training datasets with pixel-level labels is prohibitive.


Most recent works \cite{alonso2021semi, chen2021semisupervised, chen2021semi, french2020semi, hu2021semi, ouali2020semi, zhong2021pixel, wang2022semi, guan2022unbiased, liu2022perturbed, kwon2022semi, yang2022st++, zhao2022augmentation} belong to the deterministic approach that aims at directly making a prediction for an input image. That is, it does not model the predictive distribution, and only gives a point estimate. In contrast,  a method modeling a predictive distribution for the input is classified as probabilistic approach. Its key advantage is that one can estimate the uncertainty for an input by simply sampling from the posterior. 
The uncertainty provides information about whether the prediction is reliable, and thus how to estimate uncertainty should be considered under the setting of semi-supervised learning (SSL), as the performance of segmentation models is vulnerable to unlabeled data with incorrect pseudo-labels, and decision-makers need to know when they should not trust the models.

Unfortunately, the probabilistic approach is insufficiently investigated, 
as researchers barely explored its application to  semi-supervised semantic segmentation for computer vision, and most related works focus on medical imaging \cite{sedai2019uncertainty, shi2021inconsistency, yu2019uncertainty, li2020self, wang2021tripled, wang2022rethinking, meyer2021uncertainty, xiang2022fussnet}, in which Monte Carlo (MC) dropout has been the mainstream option for uncertainty quantification. 
MC dropout, however, has some limitations that hinder it from real scenarios. For instance, it can be time-consuming when it is combined with cumbersome segmentation models, as several feedforward passes are required for estimating uncertainty. In addition, architectural choices, such as where to insert dropout layers and the value of the dropout rate, are usually empirically set, which may result in a suboptimal performance. To tackle the limitations, a very recent work \cite{wang2022np} has studied neural processes (NPs)
for SSL, in which a new model named NP-Match is proposed.   
Compared to MC-dropout-based SSL models, NP-Match is computationally significantly more efficient, as it only needs to perform one feedforward pass to derive the prediction with an uncertainty estimate for a given input. 
Moreover, in NP-Match, NPs are directly built on top of convolutional neural networks (CNNs), and hence, unlike MC dropout, which has to be empirically set, NPs are more convenient to use. 
 
Considering the success of NP-Match and insufficient exploration towards the  probabilistic approach for semi-supervised semantic segmentation, in this work, we investigate the application of NPs on  semi-supervised semantic segmentation, and propose a new model, called NP-SemiSeg. 
In particular, we primarily made two modifications when designing NP-SemiSeg. First, a global latent variable is predicted for each input image, rather than producing a global latent vector shared by different images.\footnote{In NP-Match \protect\cite{wang2022np}, NPs generate a global latent vector shared by all images within a given batch, which follows the pipeline of the original NPs \protect\cite{garnelo2018neural}.} This change is inspired by the fact that different images may have different prior label distributions. 
Hence, it is more reasonable to assume that every image has its own specific prior, and NP-SemiSeg should separately predict a global latent vector for every image, shared by all its pixels. Second, attention mechanisms are additionally introduced
to both the deterministic path and the latent path. In the original NPs \cite{garnelo2018neural}, the information of context points or target points is summarized via a mean aggregator in both paths, and NP-Match also follows this practice. However, the mean aggregator introduces the issue that the decoder of NPs cannot capture relevant information for a given target prediction, as the mean aggregator gives the same weight to each point. Inspired by another model named attentive NPs \cite{kim2019attentive}, 
attention mechanisms are also integrated into NP-SemiSeg to solve this issue.

To validate the effectiveness of NP-SemiSeg, we conducted several experiments on two public benchmarks, namely,  PASCAL VOC 2012
and Cityscapes, with diverse SSL settings, and the results show two merits of NP-SemiSeg.  
First, NP-SemiSeg is versatile and flexible, because it can be integrated into different segmentation frameworks, such as CPS \cite{chen2021semi} or U$^2$PL \cite{wang2022semi}. Equipped with NP-SemiSeg, those frameworks are turned into probabilistic models, which are able to make predictions and quantify the uncertainty for  input samples. Second, compared to the widely used MC-dropout-based segmentation models, the segmentation models with NP-SemiSeg are faster in terms of uncertainty quantification and are able to give higher-quality uncertainty estimates with less performance degradation, indicating that NP-SemiSeg can be a good alternative probabilistic method to MC dropout.

It should be noted that the principal objective of this research is not to introduce a new segmentation approach that surpasses all state-of-the-art methods. Rather, the aim is to present a novel probabilistic model for semi-supervised semantic segmentation, capable of delivering both a good performance and reliable uncertainty estimates. Summarizing, the main contributions of this paper are:\vspace*{-2ex}
\begin{itemize}[leftmargin=*, itemsep=0.5pt]
\item We adjust NPs to semi-supervised semantic segmentation, and propose a new probabilistic model, named NP-SemiSeg, which is flexible and can be combined with different existing segmentation frameworks for making predictions and estimating uncertainty. 

\item We integrate an attention aggregator into NP-SemiSeg, which assigns higher weights to the information that is more relevant to target data, enhancing the performance of NP-SemiSeg.

\item Compared to MC-dropout-based segmentation models, NP-SemiSeg not only performs better in terms of accuracy, but also runs faster regarding uncertainty estimation, showing its potential to be a new probabilistic model for semi-supervised semantic segmentation.
 
\end{itemize}

The rest of this paper is organized as follows. In Section~\ref{sec:related_work}, we briefly discuss related works. Section~\ref{sec:methodology} elaborates our NP-SemiSeg, followed by our experimental details and results in Section~\ref{sec:experiments}. Finally, we give a conclusion and some future research directions in Section~\ref{sec:conclusion}. The source code is available at: \href{https://github.com/Jianf-Wang/NP-SemiSeg}{https://github.com/Jianf-Wang/NP-SemiSeg}.

\section{Related Works}
\label{sec:related_work}

In this section, we briefly review related works, including SSL for image classification, semi-supervised semantic segmentation, and the neural process (NP) family.

{\bf SSL for Image Classification.} In the past few years, many methods have been proposed for semi-supervised image classification, which provide insights and research directions for semi-supervised semantic segmentation. The most prevalent method is FixMatch \cite{sohn2020fixmatch}. During training, it produces pseudo-labels for weakly-augmented unlabeled data based on a preset confidence threshold, and the pseudo-labels are used as the ground-truth for their strongly augmented version to train the whole framework. FixMatch \cite{sohn2020fixmatch} thereafter inspired a series of promising methods \cite{li2021comatch, rizve2021defense, zhang2021flexmatch, nassar2021all, pham2021meta, hu2021simple}. 
For example, \citet{li2021comatch} incorporate contrastive learning through additionally designing the projection head that generates low-dimensional embeddings for samples. The low-dimensional embeddings  with similar pseudo-labels are encouraged to be close, which improves the quality of pseudo-labels. \citet{zhang2021flexmatch} use dynamic confidence thresholds 
that are adjusted based on the model's learning status of each class, rather than the fixed preset confidence threshold. A more relevant method, named uncertainty-aware pseudo-label selection (UPS) framework, was proposed by \citet{rizve2021defense}. This framework can be regarded as a probabilistic approach, as it applies MC dropout to obtain uncertainty estimates, based on which unreliable pseudo-labels are filtered out. Due to the weaknesses of MC dropout mentioned above, \citet{wang2022np} try to explore a new alternative probabilistic model, i.e., NPs, for semi-supervised image classification, and propose a new method called NP-Match, which not only shows a promising accuracy on several public benchmarks, but also alleviates the problem of MC dropout. These results encourage us to further investigate the application of NPs on semi-supervised semantic segmentation.

{\bf Semi-supervised Semantic Segmentation.} Most methods can be classified into two training paradigms, namely, consistency-training \cite{french2020semi, zhou2021c3, ouali2020semi, zhong2021pixel, liu2022perturbed, ke2019dual} and self-training \cite{alonso2021semi, chen2021semisupervised, hu2021semi, wang2022semi, guan2022unbiased, kwon2022semi, yang2022st++, zou2020pseudoseg, zhao2022augmentation}. 

The consistency-training methods aim to maintain the consistency among the segmentation results of different perturbations of the same unlabeled sample. For example, \citet{ouali2020semi} propose a cross-consistency training (CCT) method, and it contains a main decoder and several auxiliary decoders, which share the same encoder. For the unlabeled examples, a consistency between the main decoder’s outputs and the auxiliary outputs is maintained, over different kinds of perturbations leveraged
to the inputs of the auxiliary decoders. \citet{zhong2021pixel} design a new framework, named PC$^2$Seg, which takes advantage of both the pixel-contrastive property and the consistency property during training, and their combination further enhances the performance.
Considering the potential inaccurate training signal caused by perturbations, 
\citet{liu2022perturbed} introduce an additional teacher model, a stricter confidence-weighted cross-entropy loss, and a new type of feature perturbation to improve  consistency learning.  

Self-training methods assign pixel-wise pseudo-labels to unlabeled data, and re-train the segmentation networks. For instance,  PseudoSeg \cite{zou2020pseudoseg} utilizes the predictions of unlabeled data as the labels to re-train the whole framework. To obtain accurate pseudo-labels, a calibrated fusion module is incorporated, which fuses both the outputs of the decoder and the refined class activation map (CAM). The success of self-supervised learning motivates \citet{alonso2021semi} to integrate the pixel-level contrastive learning scheme into their framework, which aims at enforcing the feature vector of a target pixel to be similar to the same-class features from the memory bank. Recently, \citet{wang2022semi} have discovered that some pixels may never be learned in the entire self-training process, due to their low confidence scores. Then, they propose a new framework,  called U$^2$PL, which reconsiders those pixels as negative samples for training. \citet{zhao2022augmentation} reconsider the data augmentation techniques used in the self-training process, and they design a new highly random intensity-based augmentation method and an adaptive cutmix-based augmentation method to enhance the performance. 

All above methods do not involve any probabilistic model, and it is only valued in medical imaging \cite{sedai2019uncertainty, shi2021inconsistency, yu2019uncertainty, li2020self, wang2021tripled, wang2022rethinking, meyer2021uncertainty, xiang2022fussnet}, where most methods rely on MC dropout for approximating BNNs and estimating uncertainty. In a nutshell, those methods usually leverage uncertainty maps given by MC dropout to refine pseudo-labels for unlabeled data, thereby boosting the capability of their models.

{\bf NP Family.}  The first member of the NP family comes from \citet{garnelo2018conditional}; it is called conditional NP (CNP).  CNPs model the predictive distribution over context sets and target sets. However, CNPs only provide a point-wise uncertainty estimate. In most cases, it would be beneficial to exploit the correlation among different points during inference. Therefore, NPs are proposed to build the correlation points by introducing global latent variables as priors for those points \cite{garnelo2018neural}. \citet{kim2019attentive} have observed that NPs tend to underfit the context set, which is caused by the mean aggregator giving equal weights to all the context points. To remedy this issue, they propose a new model, called attentive NP, which uses an attention mechanism to attend to relevant context points with respect to target predictions. Concerning that the application areas of NPs are time series or spatial data, the translation equivalence should be an important property, i.e., if the data are translated in time or space, the predictions should be translated correspondingly. This property was ignored in previous models, until \citet{gordon2019convolutional} designed a new model called convolutional CNPs. Besides, concerning that the global latent variables are not flexible for encoding inductive biases, \citet{louizos2019functional} employ local
latent variables along with a dependency structure among them instead, obtaining a new functional NP (FNP). Similarly, \citet{lee2020bootstrapping} also point out the limited flexibility of a single latent variable to model functional uncertainty, and they use a classic frequentist technique, namely,  bootstrapping, to model functional uncertainty, leading to a new NP variant, named Bootstrapping Neural Processes (BNPs). \citet{bruinsma2021gaussian} propose a new NP variant called Gaussian NPs (GNPs), which not only involves translation
equivariance with Gaussian processes \cite{gpml}, but also provides universal approximation guarantees. Note that we only summarize some classical members of the NP family  in this part, and for more variants and their applications, please refer to the survey paper \cite{jha2022neural}.

\section{Methodology}
\label{sec:methodology}

\subsection{Background}
 
Neural Processes (NPs) are a neural network-based formulation that learn to approximate a stochastic process through finite-dimensional marginal distributions \cite{garnelo2018neural}. Their working mechanism is closely related to a classical non-parametric model, Gaussian Processes (GPs). A GP makes the assumption that each point in the input space essentially maps to a normally distributed random variable. The GP model is fully specified by a mean function, which provides the expected value of these random variables, and a kernel function, which describes the dependencies among the variables. Thus, GPs are a powerful probabilistic model that can provide a measure of uncertainty along with predictions. However, GPs are computationally expensive for large datasets and require a careful choice and tuning of the kernel function, which hinders their practical applications. To address these issues, NPs have been proposed.
 
 Before formally defining NPs, we first give the definition of a stochastic process. 
 In general, a stochastic process can be  defined as $\{F(x, \omega) : x \in \mathcal{X} \}$ over a probability space $(\Omega, \Sigma, \Pi)$ and an index set $\mathcal{X}$, 
where $F(\cdot\ , \ \omega)$ is a sample function
mapping $\mathcal{X}$ to another space $\mathcal{Y}$ for any point $\omega \in \Omega$. Therefore, for any finite sequence $x_{1:n}$, a marginal joint distribution function can be defined on the function values $F(x_1, \ \omega), F(x_2, \ \omega), \ldots, F(x_n, \ \omega)$, which satisfies two conditions given by the Kolmogorov Extension Theorem \cite{oksendal2003stochastic}: 
 \vspace{1ex}
 
{\bf Exchangeability}: \emph{This condition indicates that the marginal joint distribution should remain unaffected by any permutation of the sequence.}
 \vspace{1ex}
 
{\bf Consistency}: \emph{ This condition requires that the marginal joint distribution should remain unaffected when a part of the sequence is marginalized out. }
 \vspace{1ex}
 
With the two conditions, a stochastic process can be described by the marginal joint distribution function, namely: 
\begin{equation}
\small
\label{eq:joint1}
p(y_{1:n}|x_{1:n}) = \int \pi(\omega) p(y_{1:n}|F(\cdot\ , \ \omega), x_{1:n}) d\mu(\omega),
\end{equation}
where $\pi$ denotes density, namely,  $d\Pi = \pi d\mu$. Here, the function $F(\cdot\ , \ \omega)$ is determined by the kernels, which measure how all variables interact with each other.

To approximate stochastic processes, NPs parameterize the function $F(\cdot\ , \ \omega)$ in the marginal joint distribution with neural networks and latent vectors.  Specifically, let $(\Omega, \Sigma)$ be $(\mathbb{R}^d, \mathcal{B}(\mathbb{R}^d))$,  where $\mathcal{B}(\mathbb{R}^d)$ denotes the {\it Borel} $\sigma${\it -algebra} of $\mathbb{R}^d$, and NPs use a latent vector $z \in \mathbb{R}^d$ sampled from a multivariate Gaussian distribution to govern the function $F(\cdot\ , \ \omega)$. Then, $F(x_i , \ \omega)$ can be replaced by $\phi(x_i, z)$, where $\phi(\cdot)$ denotes a neural network, and Eq.~(\ref{eq:joint1}) becomes:
\begin{equation}
\small
\label{eq:joint2}
p(y_{1:n}|x_{1:n}) = \int \pi(z)p(y_{1:n}|\phi(x_{1:n}, z), x_{1:n}) d\mu(z).
\end{equation} 
By doing this, NPs are capable of predicting and estimating uncertainty for each data point, circumventing the explicit access to kernel functions and comparisons of distances among distinct points. This capability renders them practical for application in real-world scenarios.

The training objective of NPs is to maximize $p(y_{1:n}|x_{1:n})$, which can be implemented by  maximizing its evidence lower-bound (ELBO). The learning procedure reflects the NPs' property that they have the capability to make predictions for target data conditioned on context data \cite{garnelo2018neural}.

\subsection{NP-SemiSeg}

\subsubsection{NPs for semi-supervised semantic segmentation}

Semantic segmentation can be treated as a pixel-wise classification problem, and therefore, $p(y_{1:n}|\phi(x_{1:n}, z), x_{1:n})$ in Eq.~(\ref{eq:joint2}) can be changed to the categorical distribution (denoted as $\mathcal{C}$). 
Specifically, a weight matrix ($\mathcal{W}$) and a softmax function ($\Phi$) can be sequentially applied to the feature presentation of every pixel from the decoder $\phi(\cdot)$, outputting a probability vector that can parameterize  $\mathcal{C}$. 
Furthermore, different images can have distinct prior label distributions, as some objects cannot appear in the same image. For example, if an image captures the main road of a city, fish will not appear, whose prior should be zero. But if the image records the creatures in the sea, the prior of fish is close to one.  
Because of this, rather than using a global latent variable for different images, we instead use a latent variable per image. This can be viewed as giving each image its own prior. 
Thus, we rewrite $p(y_{1:n}|\phi(x_{1:n}, z), x_{1:n})$ as follows:
\begin{equation}
\label{eq:likelihood}
p(y_{1:n}|\phi(x_{1:n}, z_{1:n}), x_{1:n}) =  \mathcal{C}(\Phi(\mathcal{W}\phi(x_{1:n}, z_{1:n}))),
\end{equation}
where the decoder $\phi(\cdot)$ can be learned through amortised variational inference. Specifically, as for a finite sequence with length $n$, we assume  $m$ context data ($x_{1:m}$) and $r$ target data ($x_{m+1:\ m+r}$) in it, i.e., $m+r=n$. We also assume a variational distribution over latent variables, and the ELBO is given by  (with proof in the appendix):
\begin{equation}
\small
\begin{aligned}
\label{eq:elbo}
&log\ p(y_{1:n}|x_{1:n}) \ge \\ 
& \mathbb{E}_{q(z_{m+1:\ m+r}|x_{m+1:\ m+r}, y_{m+1:\ m+r})}\Big[\sum^{m+r}_{i=m+1}log\ p(y_i|z_i, x_i) - \\
&log\ \frac{q(z_{m+1:\ m+r}|x_{m+1:\ m+r}, y_{m+1:\ m+r})}{q(z_{m+1:\ m+r}|x_{1:m}, y_{1:m})}\Big] + log\ p(y_{1:m}|x_{1:m}).
\end{aligned}
\end{equation}
Then, one can maximize the ELBO to learn the NP model. During training, we follow the setting of NP-Match \cite{wang2022np} which treats only labeled data as context data and treats either labeled or unlabeled data as target data.
 
\subsubsection{NP-SemiSeg Pipeline} 
We formulate NP-SemiSeg in a modular fashion, so that it can  directly replace the classification layer of any segmentation pipeline without changing other modules in the pipeline, to output predictions with uncertainty estimates. 
As a result, NP-SemiSeg is flexible and can be used for different segmentation frameworks. To achieve this goal, the input of NP-SemiSeg should be feature maps,\footnote{In general, most segmentation frameworks are based on DeepLab \cite{chen2017deeplab}, where a classifier acts on the final output feature maps from the decoder to predict for every location.} which is consistent with the input of a classifier in other segmentation frameworks. To make explanations clearer, we only focus on NP-SemiSeg itself.

The overall pipeline of NP-SemiSeg is shown in Figure~\ref{fig:npsemiseg}, where we represent the context and target data as generic feature maps, which could be obtained from any semantic segmentation pipeline such as U$^2$PL \cite{wang2022semi} and AugSeg \cite{zhao2022augmentation}. NP-SemiSeg has a training mode and an inference mode. The former aims to calculate loss functions with real labels or pseudo-labels during training, while the latter makes predictions for unlabeled data during training or test data during testing.
In what follows, we describe these two modes:

\begin{figure*}[t]
\centering
\includegraphics[width=\linewidth]{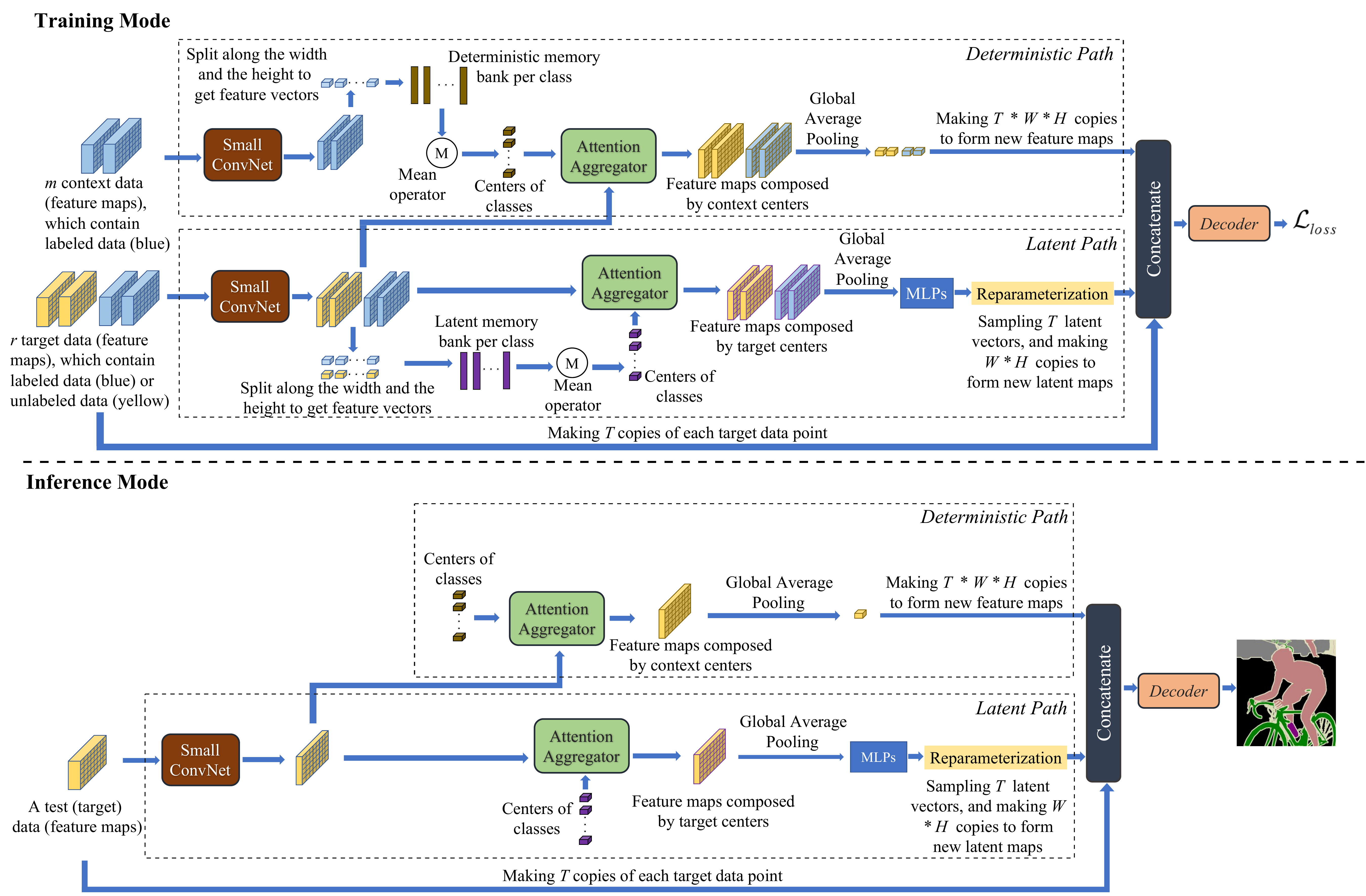}
\vspace{-3.5ex}
\caption{
Overview of NP-SemiSeg. Both the small ConvNet and the attention aggregator are shared by the deterministic path and the latent path. $T$, $W$, and $H$ represent the number of sampled latent vectors, and the width and height of the input feature maps, respectively.}
\vspace{-1ex}
\label{fig:npsemiseg} 
\end{figure*}

{\bf Training mode.} 
Given a batch of labeled data and a batch of unlabeled data, NP-SemiSeg is initially switched to inference mode, and it makes predictions for the unlabeled data. Those predictions are regarded as pseudo-labels for unlabeled data by taking the class with the highest probability. Then, NP-SemiSeg  turns to training mode, and it duplicates the labeled samples and treats them as context data. 
Subsequently, the context data are passed through a deterministic path, which aims to obtain order-invariant context representations, and the target data are passed through a latent path,  which aims to produce latent variables. The outputs from both paths are finally concatenated and then passed through a decoder before the loss is computed. Below, we provide details for the latent path and the deterministic path.

As for the latent path, target data are processed by a small ConvNet\footnote{The small ConvNet is mainly composed by $1\times1$ convolutions, and its outputs have the same spatial size as its inputs.} at first for dimensionality reduction, whose outputs are transformed feature maps with a low channel dimension. The transformed feature maps are further split along the width ($W$) and the height ($H$), resulting in feature vectors. Based on the number of classes, a set of latent memory banks have been initialized, each of which is assigned to a category. Those feature vectors are passed to the latent memory banks according to their real or pseudo labels.\footnote{Note that $q(z_*|x_{m+1:m+r}, y_{m+1:m+r})$ is conditioned on both data and labels, which is implemented by using them as inputs to MLPs in NP-Match, but in NP-SemiSeg, the condition on labels is implicitly implemented, i.e., how data are stored in memory banks is  determined by the labels.} Then, a mean operator is used for each memory bank, and we can obtain a center for each class. 
Those centers and the target transformed feature maps are input to an attention aggregator, whose outputs are feature maps composed by target centers. 
Specifically, 
the feature vector of each location in such centers-based feature maps is the weighted summation of the centers, which intends to represent every location by the most relevant features from the memory. 
Thereafter, the global average pooling and MLPs are used to produce a mean vector and a variance vector for each target data point, followed by a reparameterization trick to get $T$ latent vectors whose dimension is $D_t$. Finally, those latent vectors are copied for $W \times H$ times, thereby forming latent maps for each target data point with size $T \times D_t \times W \times H$. 

As for the deterministic path, context data are processed in the same way as the target data, until we obtain the context centers for classes. Then, the context centers as well as the  target transformed feature maps are fed to the attention aggregator, in order to get the feature maps composed by the context centers, which are further processed by global average pooling, leading to an order-invariant context representation with dimension $D_c$ for each target data point. Finally, the order-invariant context representation is copied for $T \times W \times H$ times, thereby forming context maps for each target data point with size $T \times D_c \times W \times H$. 

After the latent maps and the context maps are obtained for each target data point, they are concatenated with the original feature maps of the target data whose size is $T \times D \times W \times H$, and the concatenated feature maps will have the size $T \times (D + D_t + D_c) \times W \times H$, based on which a decoder $\phi(\cdot)$ makes pixel-wise predictions. The final prediction for each target data point can be obtained by averaging the $T$ prediction maps, and the uncertainty map is computed as the entropy of the average prediction \cite{kendall2017uncertainties}. For saving space, only those centers are stored for inference after training, instead of saving those memory banks.

{\bf  Inference mode.} As for a set of test data, they are treated as target data and are first processed by the small ConvNet. Its outputs, the target centers, and the context centers are taken as inputs to the attention aggregator to acquire the feature maps composed by centers. Subsequently, the remaining steps are the same as in the training mode to generate concatenated feature maps where the decoder $\phi(\cdot)$ acts on to make predictions, along with their associated uncertainty estimates.

\subsubsection{Attention Aggregator}
To predict a target data point, it is beneficial to gather relevant information from memory banks, as the centers close to the target provide similar representations. To achieve this, an attention aggregator is required, whose role is to produce centers-based feature maps based on the distance between query feature maps and a set of centers.   We denote the input feature maps and the input centers as $\mathcal{M}$ and $\mathbf{C}$, respectively.  The output $\mathcal{M}_\mathbf{C}$ is calculated as follows:
\begin{equation}
\small
\begin{aligned}
\label{eq:atten}
 \mathcal{M}_\mathbf{C}[i, j] = \sum_l \frac{e^{-\Theta(\mathcal{M}[i, j], \mathbf{C}[l]))}}{\sum_k e^{-\Theta(\mathcal{M}[i, j], \mathbf{C}[k]))}} \mathbf{C}[l],
\end{aligned}
\end{equation}
where $i$ and $j$ denote the index of feature maps along width and height. Both $l$ and $k$ denote the index of centers. $\Theta$ is defined as Euclidean distance over two vectors. In summary, the attention aggregator uses $\Theta$ to calculate the distance between the feature vector $\mathcal{M}[i, j]$ at the location ($i$, $j$) and every center, and all distances are further used to calculate weights through the softmax function for centers. Then, the output feature at location [$i$, $j$], namely,  $\mathcal{M}_\mathbf{C}[i, j]$, is the weighted combination of those centers. 
Similarly to ANPs \cite{kim2019attentive}, by using an attention aggregator, only the relevant information from the latent path and the deterministic path is involved for making predictions, thereby improving the model's performance. 

\subsubsection{Loss Functions} 
The loss function for NP-SemiSeg is derived from the ELBO (Eq.~(\ref{eq:elbo})). In particular, the first term can be achieved by pixel-wise cross entropy loss $L_c$ for both labeled and unlabeled data, which is widely used in different segmentation frameworks. The second term is the KL divergence between $q(z_{m+1:\ m+r}|x_{m+1:\ m+r}, y_{m+1:\ m+r})$ and $q(z_{m+1:\ m+r}|x_{1:m}, y_{1:m})$. Due to the i.i.d assumption, those $z_*$ are conditionally independent, and thus they can be calculated independently. We assume that the variational distribution follows a multivariate Gaussian with independent components, and for each target sample, the KL divergence term can be analytically written as:
\begin{equation}
\small
\begin{aligned}
\label{eq:kl}
L_{kl} =& 0.5 \times [\sum_{D_t} log\frac{\sigma_c^2}{\sigma_t^2} + \sum_{D_t} \frac{\sigma_t^2}{\sigma_c^2}  -  D_t + \\
& (m_c-m_t)diag({\sigma_c^{-2}})(m_c-m_t)^T],
\end{aligned}
\end{equation}
where $diag(\cdot)$ receives a vector and converts it into a diagonal matrix. $m_c$ and $m_t$ denote the mean vector of $q(z_*|x_{1:m}, y_{1:m})$ and  $q(z_*| y_{m+1:\ m+r})$, respectively. Similarly, $\sigma_c^2$ and $\sigma_t^2$ denote the variance vector of $q(z_*|x_{1:m}, y_{1:m})$ and  $q(z_*| y_{m+1:\ m+r})$, respectively. 
The third term is a conditional distribution over the context data, but it is ignored in our loss function, as its maximization has been implicitly implemented by the attention aggregator, i.e., matching the transformed feature maps to the centers (classes) according to their distances. 
The overall loss function for NP-SemiSeg can be written as:
\begin{equation}
\small
\begin{aligned}
\label{eq:all}
L_{loss} = L_c + \lambda_{kl} L_{kl},
\end{aligned}
\end{equation}
where $\lambda_{kl}$ is the coefficient. When NP-SemiSeg is incorporated into different segmentation frameworks, $L_{loss}$ can be naturally incorporated into their loss functions for end-to-end training.

\begin{table}
\centering 
\resizebox{0.45\textwidth}{!}{
\begin{tabular}{@{}ccccc@{}}
 \toprule[1pt]
Method & 1/16 (92) &  1/8 (183)   & 1/4 (366)  & 1/2 (732)  \\
 \hline 
  MT   & 48.37  & 58.44  &  65.49  &  68.92 \\ 
  PS-MT & 63.32  & 67.78   &  74.68  &  76.54 \\ 
  U$^2$PL  &  62.13  &  68.11  &  73.22  &  75.60  \\ 
  AugSeg  & 64.22 &  72.17   &  76.17  &   77.40 \\ 
 \hline
 MT   w/ MC dropout   & 47.78  & 57.02   &  64.82  &  67.79 \\ 
 PS-MT w/ MC dropout  & 62.09  & 66.46   &  73.11 &  74.30 \\ 
 U$^2$PL w/ MC dropout & 59.17  & 66.89  & 72.16 &  74.19 \\ 
 AugSeg w/ MC dropout  & 62.78  & 69.87   & 74.76  &  76.13 \\
\hline
 MT   w/ NP-SemiSeg   &  49.02  &  58.91   &  65.27  &  69.34  \\ 
 PS-MT w/ NP-SemiSeg  & 63.76   &   68.17   &  74.93  &   76.33   \\ 
 U$^2$PL w/ NP-SemiSeg  &  59.45  &  68.73   &  74.16   &  75.77 \\ 
 AugSeg w/ NP-SemiSeg  & 65.78  &  72.38  & 75.77 &  77.40   \\ 
  \bottomrule[1pt]
  \end{tabular}}\vspace*{-1ex}
 \caption{The mean IoU of different frameworks using ResNet-50 with either MC dropout or NP-SemiSeg on the  {\it classic} PASCAL VOC 2012 validation set under different partition protocols.}  
 \label{tab:voc_classic}
 \end{table}

\begin{table}
\centering 
\resizebox{0.47\textwidth}{!}{
\begin{tabular}{@{}ccccc@{}}
 \toprule[1pt]
Method & 1/16 (662) & 1/8 (1323)  & 1/4 (2646) & 1/2 (5291)   \\
 \hline 
 MT  & 66.77  & 70.78    &  73.22  & 75.29 \\ 
 PS-MT  & 72.83   & 75.70   &  76.43 &  77.88 \\ 
 U$^2$PL  & 74.74  &  77.44  &  77.51   &  78.62  \\ 
 AugSeg  &  77.28   & 78.27  &   78.24  &  79.02 \\ 
 \hline
 MT   w/ MC dropout   &  65.46 &  69.29 &  72.39  &  74.67 \\ 
 PS-MT w/ MC dropout  & 71.28  &   74.03 &  74.97 &  75.97
 \\\ 
 U$^2$PL w/ MC dropout & 73.79  & 76.23  &  76.56  &  76.41 \\ 
 AugSeg w/ MC dropout  & 76.42  &  76.87    &  77.02   &  77.56   \\
\hline
 MT   w/ NP-SemiSeg   &  66.93   &  71.25  &  73.10  &  75.31 \\ 
 PS-MT w/ NP-SemiSeg  &  73.44  &  76.58  & 76.74  &  76.82 \\ 
 U$^2$PL w/ NP-SemiSeg  & 75.59  &  77.77 & 77.78  & 77.23  \\ 
 AugSeg w/ NP-SemiSeg  & 77.00  & 78.68  &  78.69 &  79.03 \\ 
  \bottomrule[1pt]
  \end{tabular}}\vspace*{-1ex}
 \caption{The mean IoU of different frameworks using ResNet-50 with either MC dropout or NP-SemiSeg on the {\it blender} PASCAL VOC 2012 validation set under different partition protocols.}  
 \label{tab:voc_blender} 
 \end{table}

\begin{table}
\centering 
\resizebox{0.45\textwidth}{!}{
\begin{tabular}{@{}ccccc@{}}
 \toprule[1pt]
Method & 1/16 (186) &  1/8 (372)     & 1/4 (744) &  1/2 (1488)  \\
 \hline 
 MT  & 66.14  & 72.03  &  74.47   & 77.43 \\ 
 PS-MT  & 70.12  &  74.49   &   76.12   &  77.64 \\ 
 U$^2$PL  &  69.03   & 73.02  &   76.31  &   78.64 \\ 
 AugSeg  & 73.73   & 76.49   &  78.76  & 79.33 \\ 
 \hline
 MT   w/ MC dropout   &  65.25 &  71.09 & 72.48  &  74.96 \\ 
 PS-MT w/ MC dropout  &  68.83 &  73.11  & 75.25  &  75.47  \\ 
 U$^2$PL w/ MC dropout & 67.89 &  72.13 & 75.11  &  75.85 \\ 
 AugSeg w/ MC dropout  & 72.28  &  75.84  &  77.69 & 78.04   \\
\hline
 MT   w/ NP-SemiSeg   &  66.20  & 72.14  &   73.89 &  76.29  \\ 
 PS-MT w/ NP-SemiSeg  & 70.27  &  74.67  &  76.14 & 76.93  \\ 
 U$^2$PL w/ NP-SemiSeg  & 69.10  &  73.04   &  75.79  &  75.75  \\ 
 AugSeg w/ NP-SemiSeg  &  73.01 & 77.10  & 78.82   &  78.77 \\ 
  \bottomrule[1pt]
   \end{tabular}}\vspace*{-1ex}
 \caption{The mean IoU of different frameworks using ResNet-50 with either MC dropout or NP-SemiSeg on the Cityscapes validation set under different partition protocols. }  
 \label{tab:city} 
 \end{table}

\begin{table}
\centering 
\resizebox{0.42\textwidth}{!}{
\begin{tabular}{@{}c|c|c|c@{}}
 \toprule[1pt]
Dataset & Label Amount & MC Dropout    & NP-SemiSeg \\
 \hline 
  \multirow{4}{*}{Cityscapes}   & 1/16 (186)  & 82.89   &   84.05  \\  
   & 1/8 (372)  &  82.84   &  83.97    \\ 
   & 1/4 (744)  &   82.78  &  84.55   \\ 
   & 1/2 (1488)  &  82.92 &  84.61   \\ 
  \hline 
  \multirow{4}{*}{VOC (\it classic)}   & 1/16 (92)  & 85.79  & 86.87   \\  
   & 1/8 (183)  & 86.42   &  87.98   \\ 
   & 1/4 (366)  &  87.05  &  88.74  \\ 
   & 1/2 (732)  &  87.64  &  89.69  \\ 
   \hline 
  \multirow{4}{*}{VOC (\it blender)}   & 1/16 (662)  &  88.04 &  89.62 \\  
   & 1/8 (1323)  & 87.96   &  89.87\\ 
   & 1/4 (2646)  & 88.18    &   89.99   \\ 
   & 1/2 (5291)  & 88.42  &  89.34  \\ 
  \bottomrule[1pt]
  \end{tabular}}\vspace*{-1ex}
 \caption{The PAvPU of U$^2$PL \cite{wang2022semi} using ResNet-50 with either MC dropout or NP-SemiSeg on different datasets. }  
 \label{tab:pavpu} 
 \end{table}

 \begin{figure}[t]
\centering
\includegraphics[width=\linewidth]{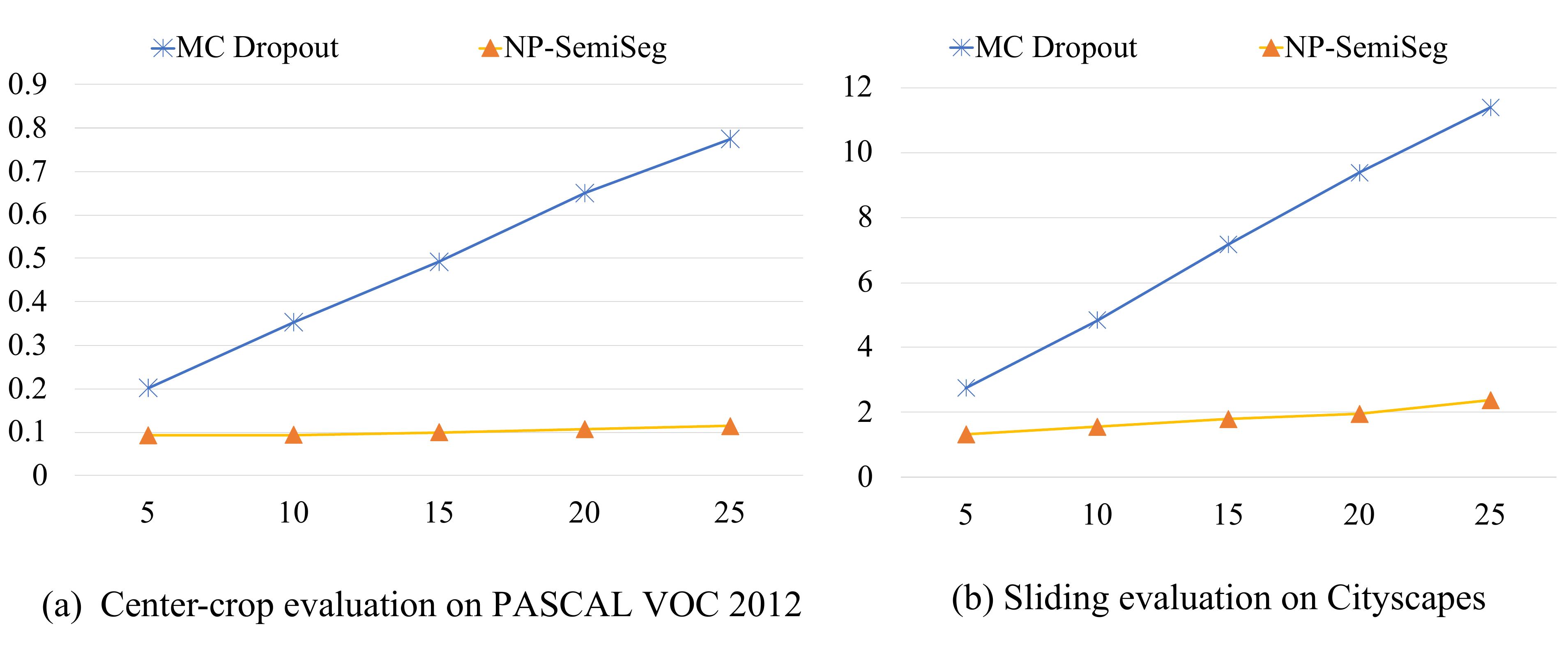}
\vspace{-5ex}
\caption{Time consumption of estimating uncertainty for U$^2$PL \cite{wang2022semi} with MC dropout and NP-SemiSeg. The horizontal axis refers to the number of predictions used for the uncertainty quantification, and the vertical axis indicates the time consumption (sec).} 
\label{fig:speed}
\end{figure}

\section{Experiments}
\label{sec:experiments}

In this section, we present our  experimental results. To save space, the 
implementation details are given in the appendix.

\subsection{Datasets}
We tested our models on two public segmentation benchmarks, namely, Cityscapes \cite{cordts2016cityscapes} and PASCAL VOC 2012 \cite{everingham2010pascal}.  Cityscapes is an urban scene understanding dataset containing 2, 975 training images with fine-annotated masks and 500 validation images. We followed previous works \cite{wang2022semi, zhao2022augmentation, chen2021semi} to use the sliding
evaluation for fair comparisons. PASCAL VOC 2012 is a standard semantic segmentation dataset that has 20 semantic classes and 1 background class. There are 1,464  and 1,449 images in the training set and the validation set, respectively. Following \citet{wang2022semi, zhao2022augmentation, chen2021semi}, we used coarsely-labeled 9,118 images from the Segmentation Boundary dataset (SBD) \cite{hariharan2011semantic}
as additional training data, and  we also evaluated our model on the {\it classic} set and the {\it blender} set. As in previous works \cite{wang2022semi, zhao2022augmentation, chen2021semi}, the center-crops of images were used for evaluation. 

\subsection{Main Results}
In the following, we report the main experimental results on the mean of Intersection over Union (mIoU),  the Patch Accuracy vs. Patch Uncertainty (PAvPU) metric \cite{mukhoti2018evaluating}, and the running time of NP-SemiSeg over the two benchmarks.

First, because of the flexibility of NP-SemiSeg, we integrated it into different segmentation frameworks to show its performance. We chose four frameworks, namely, MT \cite{tarvainen2017mean}, PS-MT \cite{liu2022perturbed}, U$^2$PL \cite{wang2022semi}, and AugSeg \cite{zhao2022augmentation}. The first two frameworks are classified as the consistency-training method, while the rest belongs to the self-training method. Since MC dropout is the mainstream probabilistic approach in SSL, we also evaluated it by applying it to the four frameworks, and it is inserted after every activation layer in their decoders.
From Tables~\ref{tab:voc_classic}, \ref{tab:voc_blender}, and \ref{tab:city}, we have two findings. First, on PASCAL VOC 2012, NP-SemiSeg can help to further improve the mIoU in most cases. In  contrast, MC dropout leads to a poor performance, and it is outperformed by NP-SemiSeg with a healthy margin. Second, on Cityscapes, though NP-SemiSeg only achieves comparable results, it still performs clearly better than MC dropout. 
Thus, compared to MC dropout, NP-SemiSeg is a more favorable choice for semi-supervised semantic segmentation, as it does not cause a serious performance degradation. 
In the other experiments, we fixed a single framework, i.e., U$^2$PL \cite{wang2022semi},  to further explore NP-SemiSeg.

\begin{table}
\centering 
\resizebox{0.45\textwidth}{!}{
\begin{tabular}{@{}c|c|c|c@{}}
 \toprule[1pt]
Dataset & Label Amount &  w/o Attention   & w/ Attention \\
 \hline 
  \multirow{4}{*}{Cityscapes}   & 1/16 (186)  & 67.86  &   69.10  \\  
   & 1/8 (372)  &  72.44  &  73.04    \\ 
   & 1/4 (744)  &  75.33  &  75.79  \\ 
   & 1/2 (1488)  &   75.45  &  75.75   \\ 
  \hline 
  \multirow{4}{*}{VOC (\it classic)}   & 1/16 (92)  &  58.52   & 59.45   \\  
   & 1/8 (183)  &  68.12  &  68.73    \\ 
   & 1/4 (366)  &   73.72 &   74.16   \\ 
   & 1/2 (732)  &  75.64  &   75.77  \\ 
   \hline 
  \multirow{4}{*}{VOC (\it blender)}   & 1/16 (662)  &   74.80  &   75.59  \\  
   & 1/8 (1323)  & 77.26  &  77.77  \\ 
   & 1/4 (2646)  & 77.38  &   77.78  \\ 
   & 1/2 (5291)  & 76.91  &  77.23   \\ 
  \bottomrule[1pt]
  \end{tabular}}\vspace*{-1ex}
 \caption{Ablation studies of attention aggregation on different datasets. The results are all based on U$^2$PL \cite{wang2022semi} using ResNet-50, and mean IoU is reported.}  
 \label{tab:ablate1} 
 \end{table}

 \begin{table}
\centering 
\resizebox{0.45\textwidth}{!}{
\begin{tabular}{@{}c|c|c|c@{}}
 \toprule[1pt]
Dataset & Label Amount &  w/o Attention   & w/ Attention \\
 \hline 
  \multirow{4}{*}{Cityscapes}   & 1/16 (186)  &  83.46  &  84.05  \\  
   & 1/8 (372)  &  83.61 &  83.97   \\ 
   & 1/4 (744)  &  84.32  &    84.55  \\ 
   & 1/2 (1488)  &  84.60  &   84.61  \\ 
  \hline 
  \multirow{4}{*}{VOC (\it classic)}   & 1/16 (92)  &  86.22  &   86.87   \\  
   & 1/8 (183)  &  87.54  &  87.98   \\ 
   & 1/4 (366)  &  88.57 &   88.74  \\ 
   & 1/2 (732)  &  89.53 &  89.69     \\ 
   \hline 
  \multirow{4}{*}{VOC (\it blender)}  
   & 1/16 (662)  &  89.46 &  89.62 \\  
   & 1/8 (1323)  &  89.53  &  89.87 \\ 
   & 1/4 (2646)  &  89.57  &   89.99   \\ 
   & 1/2 (5291)  &  89.35  &  89.34  \\ 
  \bottomrule[1pt]
  \end{tabular}}\vspace*{-1ex}
 \caption{Ablation studies of attention aggregation on different datasets. The results are all based on U$^2$PL \cite{wang2022semi} using ResNet-50, and PAvPU is reported.}  
 \label{tab:ablate2} 
 \end{table}

Second, we compare the PAvPU of NP-SemiSeg with that of MC dropout in Table~\ref{tab:pavpu} for the purpose of evaluating their uncertainty estimation. 
Under the same label amount setting for each dataset, NP-SemiSeg achieves a higher PAvPU metric than MC dropout, showing that the former
can output more reliable uncertainty estimates.
Therefore, it is more suitable than MC dropout for semi-supervised semantic segmentation in terms of uncertainty quantification.

Finally, we compare the running time of NP-SemiSeg and MC dropout for quantifying uncertainty, under two evaluation strategies, namely, the center-crop evaluation on PASCAL VOC 2012 and the sliding evaluation on Cityscapes. Note that the encoder of U$^2$PL in our experiments is a ResNet-50 \cite{he2016deep} pretrained on the ImageNet dataset \cite{deng2009imagenet}, and therefore MC dropout is only inserted into the decoder, and only the decoder performs $T$ times of feedforward passes for saving time. From Figure~\ref{fig:speed}, we have the following observations. First, when the number of predictions ($T$) increases, the time cost of MC dropout also rises accordingly, and the gap between NP-SemiSeg and MC dropout gradually becomes  significant. Second, if the sliding evaluation is used, the time consumption of MC dropout is hardly acceptable, as MC dropout requires more numbers of feedforward passes than NP-SemiSeg for this strategy. For instance, to evaluate a large image, we need to move the sliding window for $r$ strides in total, and in this case, MC dropout needs $T \times r$ feedforward passes, while NP-SemiSeg only needs $r$ feedforward passes. These observations demonstrate that NP-SemiSeg is computationally more efficient than MC dropout for semi-supervised semantic segmentation. 

\subsection{Ablation Studies}
We conducted ablation studies of the attention aggregator on two public benchmarks, which are shown in Tables~\ref{tab:ablate1} and \ref{tab:ablate2}. For the experiments without using the attention aggregator, we followed the previous work \cite{wang2022np} to use a mean aggregator for assembling the information instead. 

The results show the importance of the attention aggregator. In particular, when it is removed, we can observe that the mIoU decreases in Table~\ref{tab:ablate1}. From the perspective of uncertainty quantification, we  also see the gap regarding PAvPU between NP-SemiSeg with and without attention aggregator, even though the gap is marginal. These two findings support the significance of the attention aggregator, which involves relevant information from the memory banks to infer the latent maps and the context maps.

\section{Conclusion and Outlook}
\label{sec:conclusion}
In this work, we proposed a new probabilistic model, named NP-SemiSeg, which adjusts neural processes (NPs) to semi-supervised semantic segmentation. To better utilize the information from context data and target data, we integrated an attention aggregator into NP-SemiSeg for assigning higher weights to important information during aggregation, which is not considered in NP-Match. Our experimental results confirm the effectiveness of NP-SemiSeg in both accuracy and uncertainty estimation, thus highlighting its potential to supplant MC dropout as an innovative method for quantifying uncertainty in semi-supervised semantic segmentation.

For future research, it is valuable to explore NPs in other SSL tasks, such as object detection. In addition, it would also be interesting to see the application of NP-SemiSeg on semi-supervised medical image segmentation in the future.  

\section{Limitations}

While NP-SemiSeg is superior to MC dropout with respect to uncertainty estimation, it is important to acknowledge its performance deterioration in some SSL settings, particularly with the Cityscapes dataset. This could potentially restrict its practical application. We hypothesize two potential causes for this degradation, both of which warrant further investigation.

Firstly, during the training phase, incorrect pixel-wise pseudo-labels may be assigned to unlabeled data. This could negatively affect NP-SemiSeg's ability to approximate the variational distribution to the true distribution over latent variables, leading to a subpar performance. A similar issue in NP-Match is partially resolved through an uncertainty-guided skew-geometric Jensen-Shannon (JS) divergence. However, it is challenging to directly apply this divergence to the task of segmentation.

Secondly, considering that the performance drop is pronounced in the Cityscapes dataset, it might be attributed to the sliding evaluation strategy, which contradicts NP-SemiSeg's use of global latent variables. NP-SemiSeg operates on the premise that a single latent variable is shared among all pixels in an image. This suggests that the global latent vector is dependent on the entire content (topic) of the target image. If a sliding evaluation strategy is employed, we do not obtain a global latent vector for the entire image, but rather a latent vector for the local region covered by the sliding window. This could negatively impact performance, given the importance of global information in generating a global latent vector for a target image.

\section{Acknowledgements} 
This work was partially supported by the Alan Turing Institute under the EPSRC grant EP/N510129/1, by the AXA Research Fund, and by the EU TAILOR grant 952215.


\bibliography{example_paper}
\bibliographystyle{icml2023}

\newpage
\appendix
\onecolumn

\icmltitle{Appendix}

\section{Implementation Details}

NP-SemiSeg is a flexible module, and in our experiments, we evaluated it with four different segmentation frameworks, including MT \cite{tarvainen2017mean}, PS-MT \cite{liu2022perturbed}, U$^2$PL \cite{wang2022semi}, and AugSeg \cite{zhao2022augmentation}. When NP-SemiSeg is incorporated into them, we followed their original hyper-parameter settings for fair comparisons, and 
we only made the following changes due to limited computational resources. On the PASCAL VOC 2012 dataset, the training crop size is set to $480 \times 480$, and those frameworks with NP-SemiSeg are trained with 0.001 learning rate and 12 batch size.  On the Cityscapes dataset,  the training crop size is set to $580 \times 580$, and we used 0.005 learning rate and 8 batch size for training. When calculating PAvPU, we use a window size 64, and the uncertainty threshold is set to 0.4.  The encoder is ResNet-50 \cite{he2016deep} that is pre-trained on ImageNet \cite{deng2009imagenet}. 

The hyper-parameters of NP-SemiSeg include the length of each memory bank ($\mathcal{Q}$), 
the coefficient $\lambda_{kl}$,  the number of latent maps $T$. 
We followed NP-Match to set $\mathcal{Q}=2560$ for all memory banks. $T$ was set to 5 at both the training phase and the testing phase. The  coefficient $\lambda_{kl}$ is set to 0.005. The configuration of the small ConvNet and the decoder are separately shown in Tables~\ref{tab:small_convnet} and~\ref{tab:decoder}. The implementation of NP-SemiSeg is modified based on the public official source code of NP-Match \cite{wang2022np}. All experiments are conducted on GeForce RTX 3090 GPUs.

\begin{table}[h]
    \centering
    \resizebox{0.7\textwidth}{!}{
    \begin{tabular}{|c|c|}
       \hline
       Type  &  Configuration \\
      \hline 
      2D Conv &  \# In-C: 512, \# Out-C: 32, Kernel Size: $ 1\times1$, Stride: $ 1\times 1$, Padding:  $ 0 $ \\
      \hline
      InstanceNorm &  \# In-C: 32, \# Out-C: 32 \\
      \hline
      ReLU &  \# In-C: 32, \# Out-C: 32  \\
      \hline
      2D Conv &  \# In-C: 32, \# Out-C: 32, Kernel Size: $ 1\times1$, Stride: $ 1\times 1$, Padding:  $ 0 $ \\
      \hline
      InstanceNorm &  \# In-C: 32, \# Out-C: 32  \\
      \hline
      ReLU &  \# In-C: 32, \# Out-C: 32  \\
      \hline
      2D Conv &  \# In-C: 32, \# Out-C: 32, Kernel Size: $ 1\times1$, Stride: $ 1\times 1$, Padding:  $ 0 $ \\
      \hline
    \end{tabular}
    } 
    \caption{Configuration of the small ConvNet. It is used for dimensional reduction, in order to save GPU memory. ``In-C'' and ``Out-C'' denote the channel dimension of the input feature maps and the output feature maps, respectively.}

    \label{tab:small_convnet}
\end{table}

\begin{table}[h]
    \centering
    \resizebox{0.7\textwidth}{!}{
    \begin{tabular}{|c|c|}
       \hline
       Type  &  Configuration \\
      \hline 
      2D Conv &  \# In-C: 576, \# Out-C: 256, Kernel Size: $ 3\times3$, Stride: $ 1\times 1$, Padding:  $ 1\times 1 $ \\
      \hline
      InstanceNorm &  \# In-C: 256, \# Out-C: 256 \\
      \hline
      ReLU &   \# In-C: 256, \# Out-C: 256 \\
      \hline
      2D Conv &  \# In-C: 256, \# Out-C: 256, Kernel Size: $ 3\times3$, Stride: $ 1\times 1 $, Padding:  $ 1\times 1 $ \\
      \hline
      InstanceNorm &  \# In-C: 256, \# Out-C: 256 \\
      \hline
      ReLU &   \# In-C: 256, \# Out-C: 256 \\
      \hline
      2D Conv &  \# In-C: 256, \# Out-C: n$_{class}$, Kernel Size: $ 1\times1$, Stride: $ 1\times 1$, Padding:  $ 0 $ \\
      \hline
    \end{tabular}
    } 
    \caption{Configuration of the decoder. ``In-C'' and ``Out-C'' denote the channel dimension of the input feature maps and the output feature maps, respectively. ``n$_{class}$'' represents the number of classes.}

    \label{tab:decoder}
\end{table}

\newpage

\section{Hyper-parameter Exploration}

\begin{figure}[h]
\centering
\includegraphics[width=\linewidth]{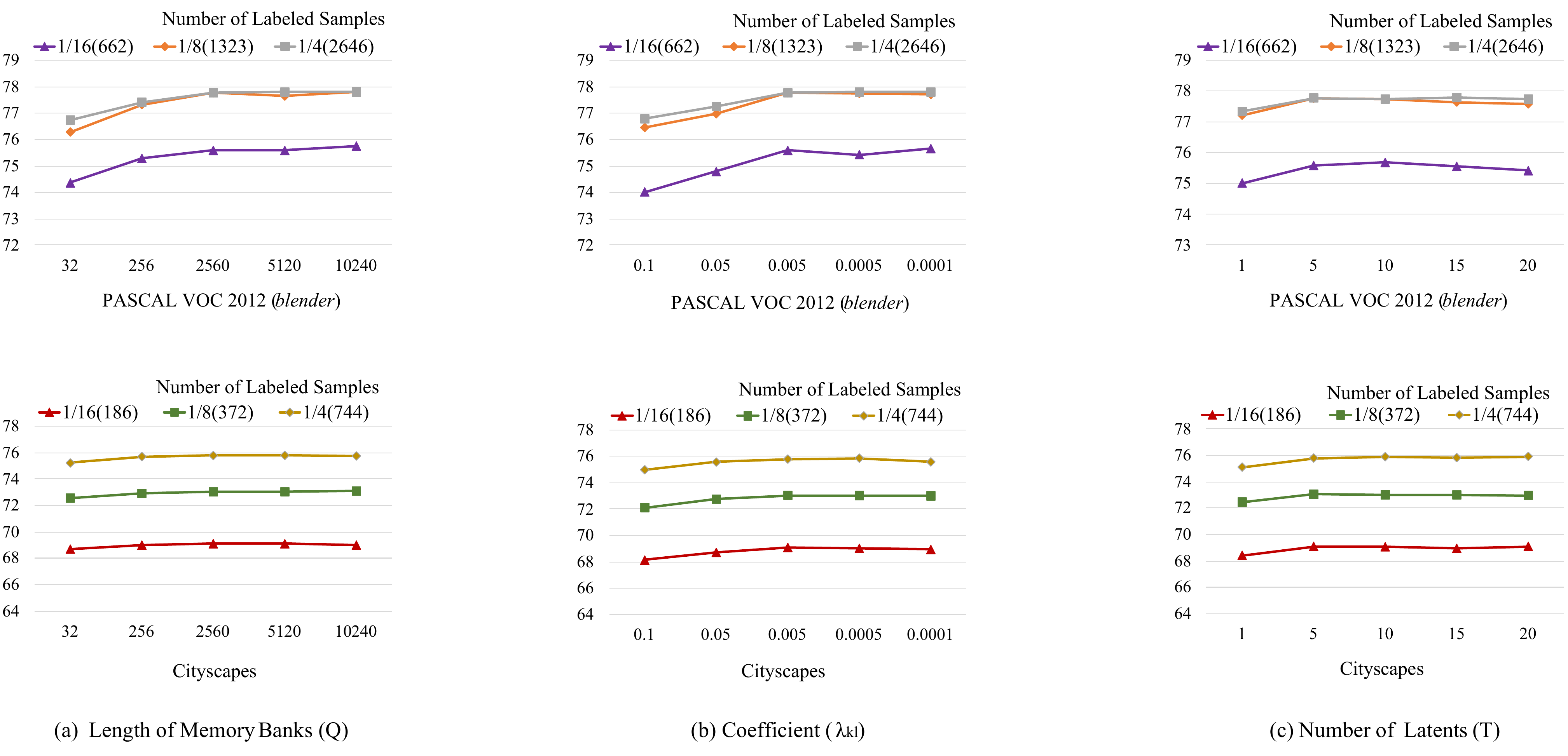}
\vspace{-4ex}
\caption{The mean IoU under different hyper-parameter settings for training.} 
\label{fig:hyper}
\end{figure}

Additional experiments are conducted on PASCAL VOC 2012 ({\it blender}) and  Cityscapes with different amounts of labeled data for hyper-parameters exploration. Three hyper-parameters are investigated in total, including the length of each memory bank ($\mathcal{Q}$),  the coefficient $\lambda_{kl}$,  and the number of latent maps $T$. By Figure~\ref{fig:hyper}(a),  $\mathcal{Q}$ should be set properly, as a small value leads to an inferior performance on both datasets. Once $\mathcal{Q}$ is large enough, further increasing the length will not affect performance. Figure~\ref{fig:hyper}(b) shows the results using different $\lambda_{kl}$. It can be seen that when $\lambda_{kl}$ rises from 0.005 to 0.1, the performance of NP-SemiSeg gets worse. Conversely, decreasing $\lambda_{kl}$ cannot impact the performance too much. Finally, Figure~\ref{fig:hyper}(c) shows the relationship between the number of latents and the performance. We can observe that the performance is insensitive to the setting of $T$, unless it is set to 1. Therefore, in our other experiments, it is a good practice to set  $\lambda_{kl}=0.005$, $\mathcal{Q}=2560$, and $T=5$.
 
\newpage
\section{Derivation of ELBO (Eq.~(\ref{eq:elbo}))}

\emph{Proof.} As for the marginal joint distribution $p(y_{1:n} | x_{1:n})$ over $n$ data points in which there are $m$ context data points and $r$ target data points (i.e., $m+r=n$), we assume a variational distribution over latent variables for the target data points, namely, $q(z_{m+1:\ m+r}|x_{m+1:\ m+r}, y_{m+1:\ m+r})$. According to the i.i.d assumption, those $z_*$ are independent from each other, and we denote its integral domain as $D_z$. Then: 
\begin{equation}
\footnotesize		
\begin{aligned} 
 &log \ p(y_{1:n} | x_{1:n}) = log \int\cdots\int_{D_z} p(z_{m+1:\ m+r}, y_{1:n} | x_{1:n}) \\
 & = log \int\cdots\int_{D_z} \frac{p(z_{m+1:\ m+r}, y_{1:n}|x_{1:n})}{q(z_{m+1:\ m+r}|x_{m+1:\ m+r}, y_{m+1:\ m+r})}q(z_{m+1:\ m+r}|x_{m+1:\ m+r}, y_{m+1:\ m+r}) \\
 & \ge \sum_{i=m+1}^{m+r}  \mathbb{E}_{q(z_i|x_{m+1:\ m+r}, y_{m+1:\ m+r})}[log \ \frac{p(z_i, y_{1:n}|x_{1:n})}{q(z_i|x_{m+1:\ m+r}, y_{m+1:\ m+r})}] \\
 & =  \mathbb{E}_{q(z_{m+1:\ m+r}|x_{m+1:\ m+r}, y_{m+1:\ m+r})}[log \ \frac{p(y_{1:m}|x_{1:m})p(z_{m+1:\ m+r}|x_{1:m}, y_{1:m})\prod^{m+r}_{i=m+1}p(y_i|z_i, x_i)}{q(z_{m+1:\ m+r}|x_{m+1:\ m+r}, y_{m+1:\ m+r})}] \\
 & = \mathbb{E}_{q(z_{m+1:\ m+r}|x_{m+1:\ m+r}, y_{m+1:\ m+r})}[\sum^{m+r}_{i=m+1} log \ p(y_i|z_i, x_i) + log \ \frac{p(z_{m+1:\ m+r}|x_{1:m}, y_{1:m})}{q(z_{m+1:\ m+r}|x_{m+1:\ m+r}, y_{m+1:\ m+r})} + log \ p(y_{1:m}|x_{1:m})] \\
 & = \mathbb{E}_{q(z_{m+1:\ m+r}|x_{m+1:\ m+r}, y_{m+1:\ m+r})}[\sum^{m+r}_{i=m+1} log \ p(y_i|z_i, x_i) - log \ \frac{q(z_{m+1:\ m+r}|x_{m+1:\ m+r}, y_{m+1:\ m+r})}{p(z_{m+1:\ m+r}|x_{1:m}, y_{1:m})}] + log \ p(y_{1:m}|x_{1:m}).
\end{aligned}
\end{equation}
Similar to NPs \cite{garnelo2018neural}, $p(z_{m+1:\ m+r}|x_{1:m}, y_{1:m})$ is unknown, we replace it with $q(z_{m+1:\ m+r}|x_{1:m}, y_{1:m})$, and then we get:
\begin{equation}
\small
\begin{aligned} 
&log\ p(y_{1:n}|x_{1:n}) \ge \\
& \mathbb{E}_{q(z_{m+1:\ m+r}|x_{m+1:\ m+r}, y_{m+1:\ m+r})}\Big[\sum^{m+r}_{i=m+1}log\ p(y_i|z_i, x_i) - log\ \frac{q(z_{m+1:\ m+r}|x_{m+1:\ m+r}, y_{m+1:\ m+r})}{q(z_{m+1:\ m+r}|x_{1:m}, y_{1:m})}\Big] + log \ p(y_{1:m}|x_{1:m}).
\end{aligned}
\end{equation}
\hfill $\square$ 

\newpage
\section{Visualization Results}

We visualize some prediction results and uncertainty maps given by NP-SemiSeg on both PASCAL VOC 2012 ({\it blender}) and Cityscapes. For the uncertainty maps, we calculate pixel-wise predictive entropy, and represent the uncertainty with gray images. 
Each uncertainty map uses pixel values, ranging from black to white, to denote the levels of uncertainty, starting from low to high.

According to the visualization results, NP-SemiSeg can provide a good quality of uncertainty estimates. In general,  it can give a high uncertainty for the pixels that are wrongly predicted. Furthermore, the boundary of an object is more likely to be misclassfied, and therefore,  NP-SemiSeg also gives  high uncertainties to boundaries. Based on this information, one can make decisions or further improve the results in a real-world scenario.

 \vspace{15ex}
\begin{figure}[h]
\centering
\includegraphics[width=\linewidth]{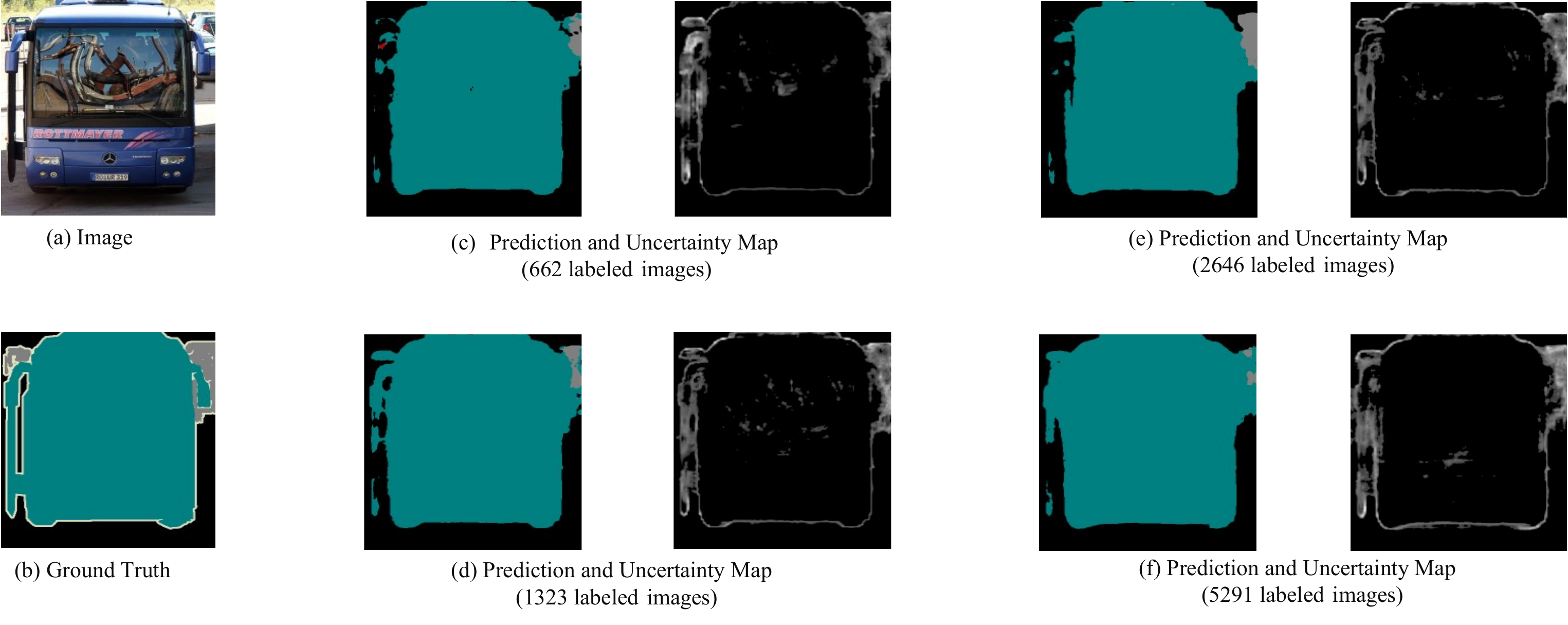}
\vspace{-2ex}
\caption{First set of visualization results on PASCAL VOC 2012 ({\it blender})  under different training protocols. The predictions and their corresponding uncertainty maps are shown. } 
\end{figure}

\clearpage

\begin{figure}[h]
\centering
\includegraphics[width= \linewidth]{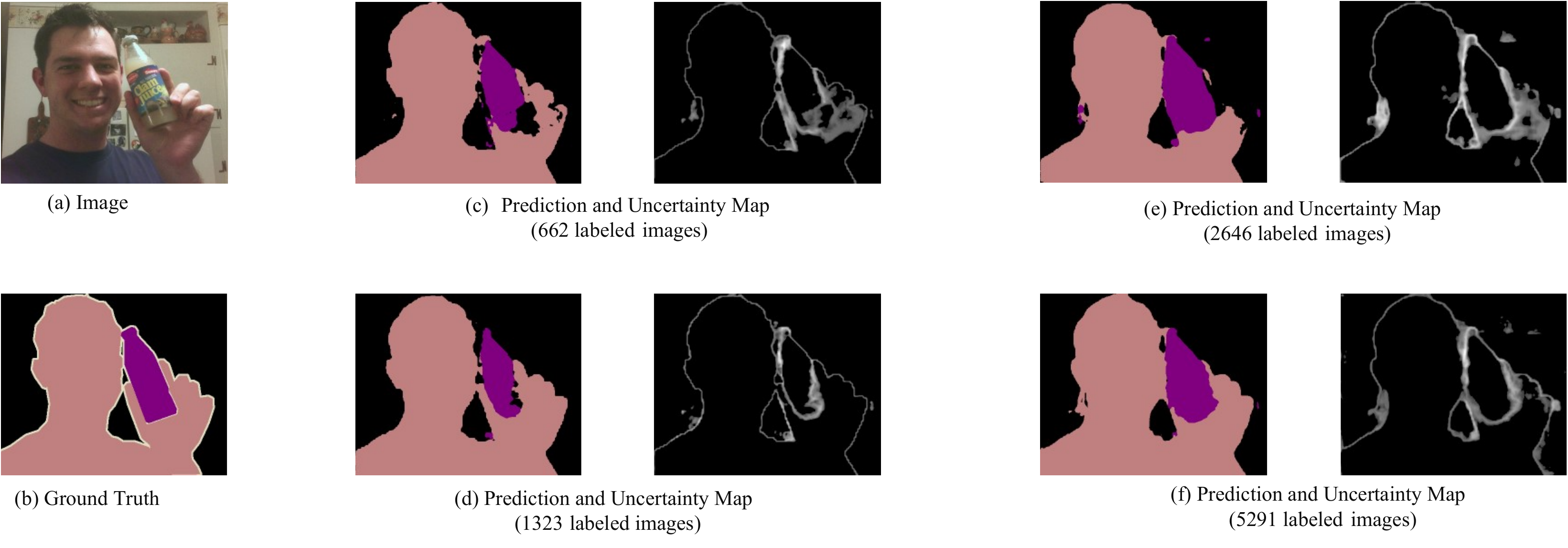}
\vspace{-2ex}
\caption{Second set of visualization results on PASCAL VOC 2012 ({\it blender})  under different training protocols. The predictions and their corresponding uncertainty maps are shown. }  
\end{figure} 

 \vspace{10ex}
 
\begin{figure}[h]
\centering
\includegraphics[width= \linewidth]{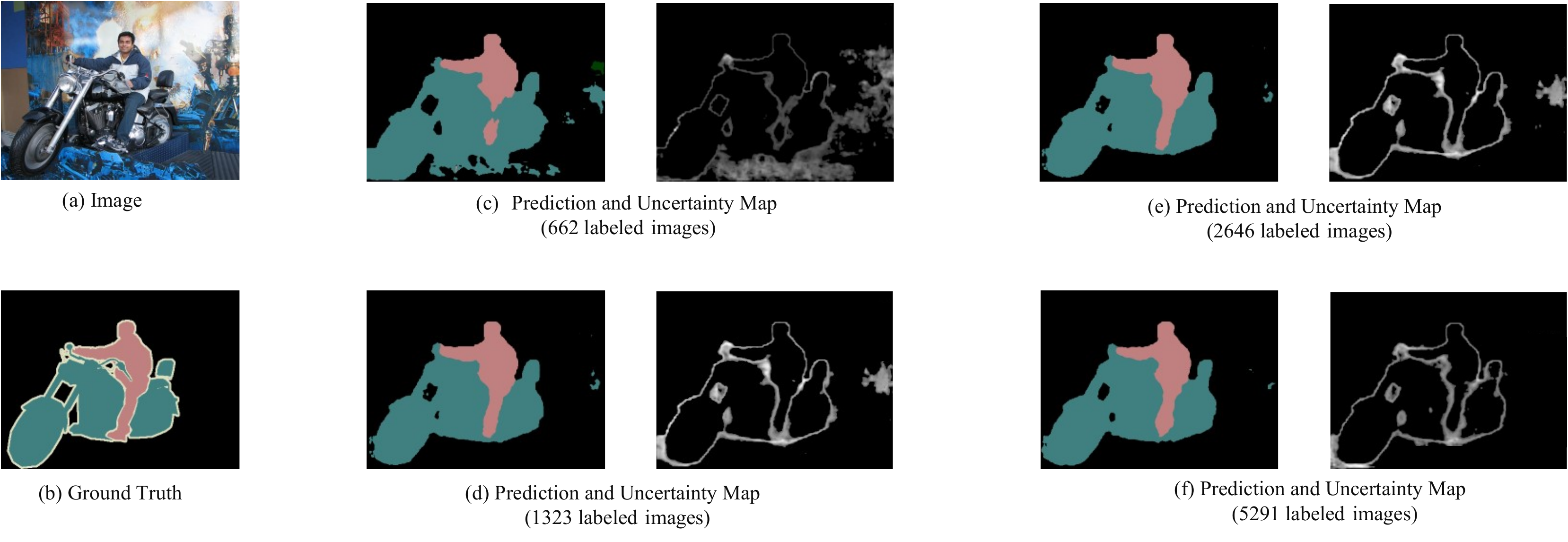}
\vspace{-2ex}
\caption{Third set of visualization results on PASCAL VOC 2012 ({\it blender})  under different training protocols. The predictions and their corresponding uncertainty maps are shown. } 
\end{figure} 
\begin{figure}[h]
\centering
\includegraphics[width= 0.95  \linewidth]{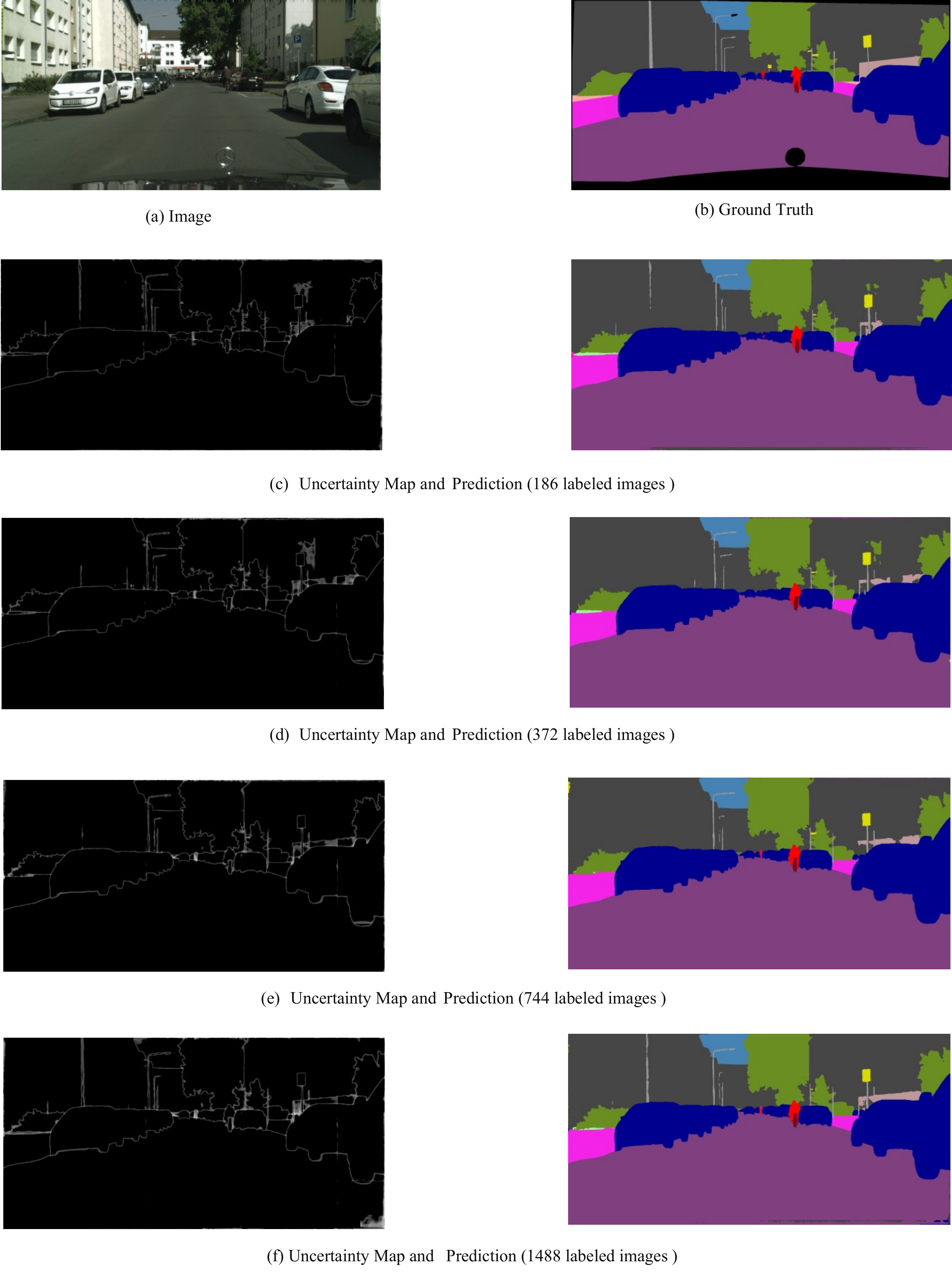}
\vspace{-2ex}
\caption{First set of visualization results on Cityscapes  under different training protocols. The predictions and their corresponding uncertainty maps are shown. } 
\end{figure} 
\begin{figure}[h]
\centering
\includegraphics[width= 0.95  \linewidth]{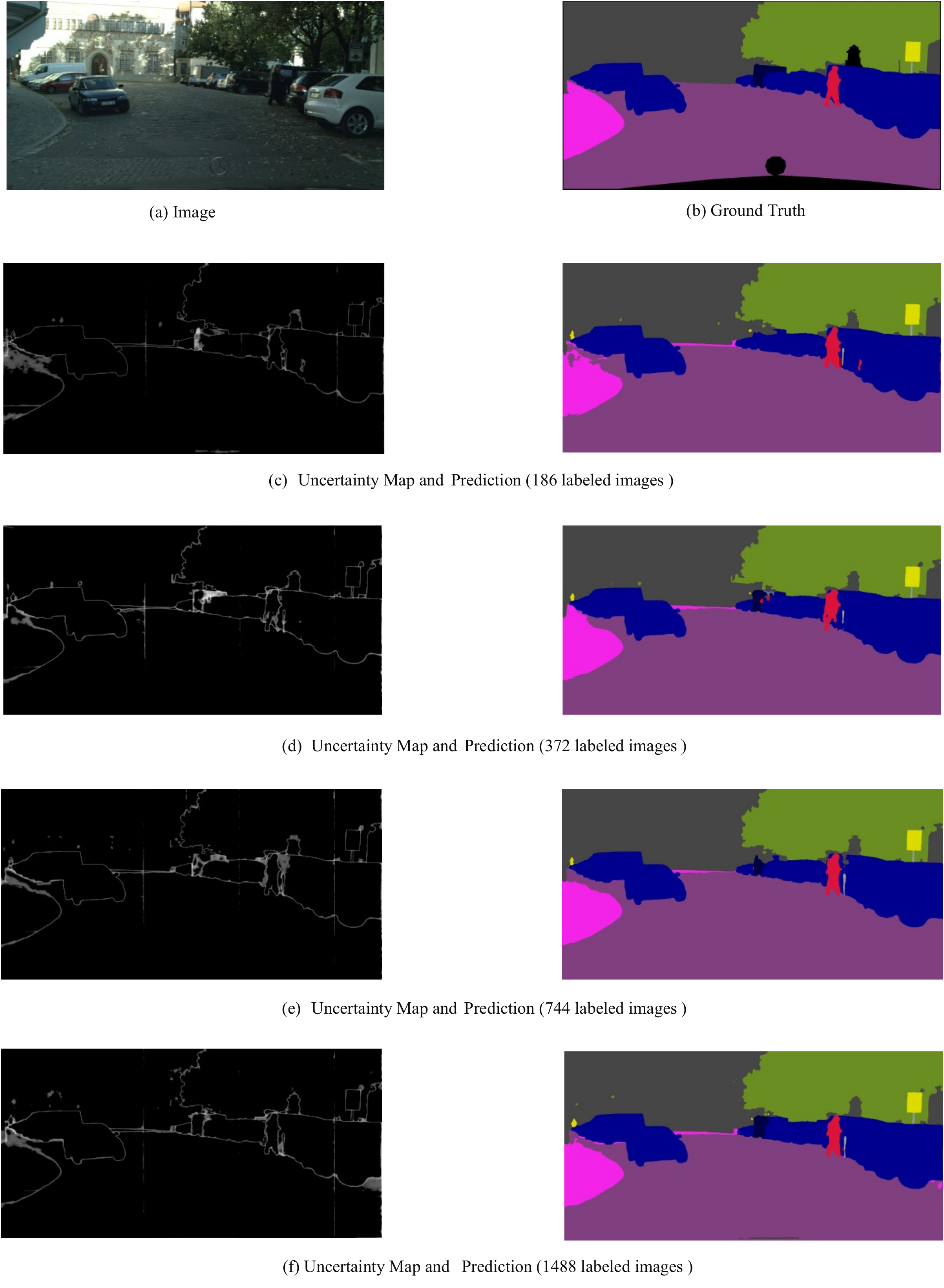}
\vspace{-2ex}
\caption{Second set of visualization results on Cityscapes under different training protocols. The predictions and their corresponding uncertainty maps are shown. }
\end{figure} 
\begin{figure}[h]
\centering
\includegraphics[width= 0.95 \linewidth]{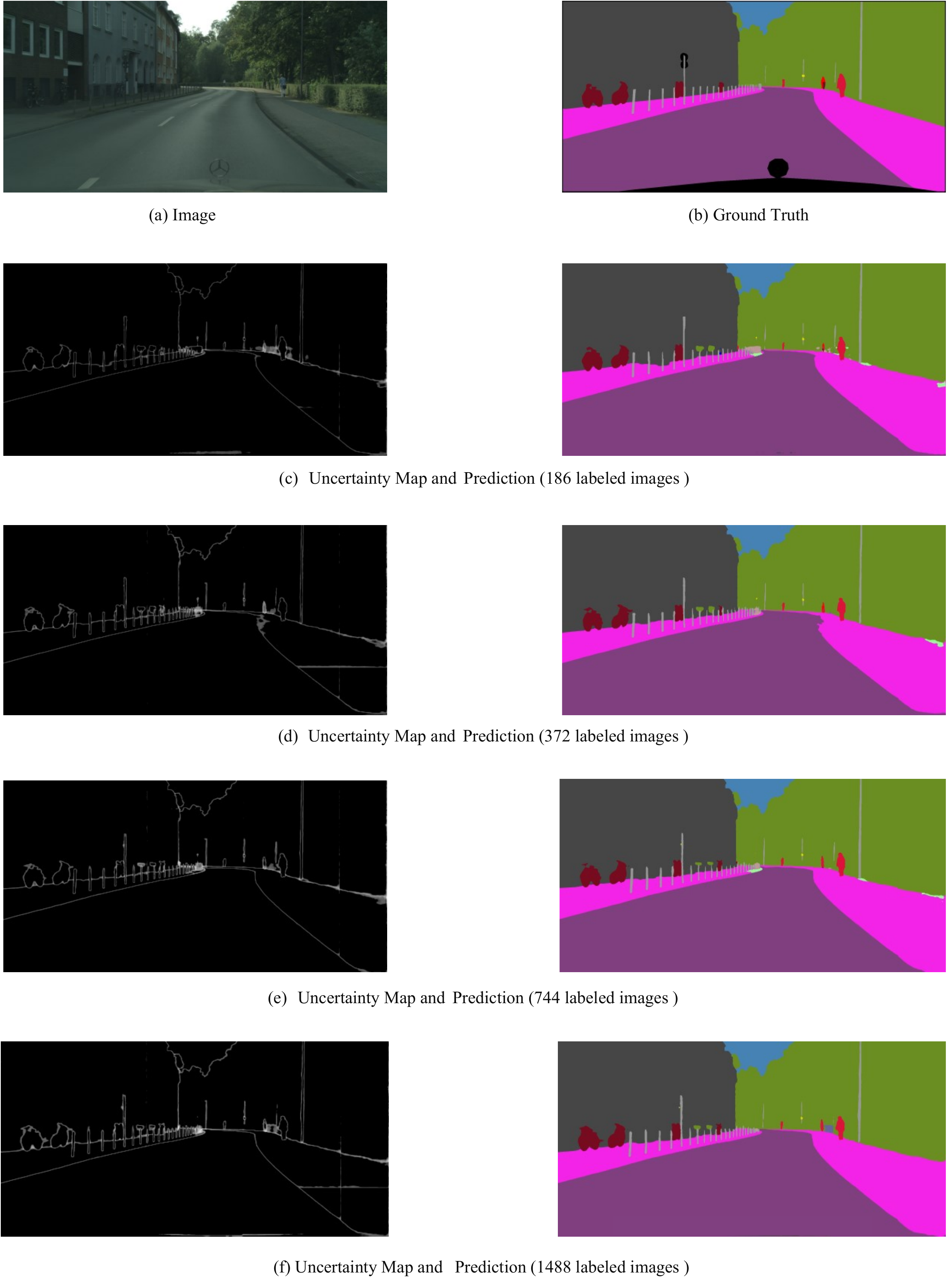}
\vspace{-2ex}
\caption{Third set of visualization results on Cityscapes  under different training protocols. The predictions and their corresponding uncertainty maps are shown. } 
\end{figure} 

\end{document}